%% file: main.tex
\pdfoutput=1

\documentclass[11pt]{article}

 \usepackage{acl}
\usepackage{times}
\usepackage{latexsym}
\usepackage{hyperref}
\usepackage{times}
\usepackage{verbatim}
\usepackage{bm}
\usepackage{algorithm}
\usepackage{algorithmic}
\usepackage{amssymb}
\usepackage{latexsym}
\usepackage{multirow}
\usepackage{float}
\usepackage{mathrsfs}
\usepackage{booktabs}
\usepackage{inconsolata}
\usepackage{graphicx}
\usepackage{multirow}
\usepackage{amsmath}
\usepackage{dsfont}
\usepackage{color}
\usepackage{enumitem}
\usepackage[T1]{fontenc}
\usepackage[T1]{fontenc}

\usepackage[utf8]{inputenc}

\usepackage{microtype}

\usepackage{inconsolata}

\title{Addressing Order Sensitivity of In-Context Demonstration Examples in Causal Language Models}

\author{{\bf Yanzheng Xiang$^{1}$ ,} {\bf Hanqi Yan$^{1}$,} {\bf Lin Gui$^1$,} {\bf Yulan He$^{1,2}$} \\
  $^1$King's College London, $^{2}$The Alan Turing Institute\\
  \texttt{\{yanzheng.xiang, hanqi.yan, lin.1.gui, yulan.he\}@kcl.ac.uk} \\
}

\begin{document}
\maketitle
\input{sections/0_abstract}
\input{sections/1_intro}

\input{sections/2_preliminary_experiments}

\input{sections/3_method}

\input{sections/4_experiment}
\input{sections/5_related_work}

\section*{Limitations}
This study primarily investigates the sensitivity of LLMs to the order of in-context examples in tasks that have definite answers. 
However, for more open-ended generation tasks, such as summarization and code generation, the impact of the order of in-context samples on the effectiveness and quality of the generated answers has yet to be explored.
We will continue to explore this type of task in the future.

\section*{Acknowledgements}
This work was supported in part by the UK Engineering and Physical Sciences Research Council (EPSRC) through a Turing AI Fellowship (grant no. EP/V020579/1, EP/V020579/2) and a New Horizons grant (grant no. EP/X019063/1).

\bibliography{custom}

\appendix
\input{sections/6_Appendix}

\end{document}

%% file: sections/0_abstract.tex
\begin{abstract}
In-context learning has become a popular paradigm in natural language processing. 
However, its performance can be significantly influenced by the order of in-context demonstration examples.
In this paper, we found that causal language models~(CausalLMs) are more sensitive to this order compared to prefix language models~(PrefixLMs).
We attribute this phenomenon to the auto-regressive attention masks within CausalLMs, which restrict each token from accessing information from subsequent tokens. 
This results in different receptive fields for samples at different positions, thereby leading to representation disparities across positions. 
To tackle this challenge, we introduce an unsupervised fine-tuning method, termed the \emph{Information-Augmented and Consistency-Enhanced} approach. This approach utilizes contrastive learning to align representations of in-context examples across different positions and introduces a consistency loss to ensure similar representations for inputs with different permutations. This enhances the model's predictive consistency across permutations.
Experimental results on five benchmarks suggest that our proposed method can reduce the sensitivity of CausalLMs to the order of in-context examples and exhibit robust generalizability, particularly when demonstrations are sourced from a candidate pool different from that used in the training phase, or when the number of in-context examples differs from what is used during training.

\end{abstract}

%% file: sections/1_intro.tex
\section{Introduction}
Large Language Models (LLMs) have demonstrated remarkable capabilities in various tasks~\citep{Wei2022ChainOT,Zhao2021CalibrateBU,Liu2021WhatMG} by being conditioned on just a few input-label pairs as demonstrations, which is referred to as ``\emph{In-context learning}~(ICL)''~\citep{Brown2020LanguageMA}.
Despite its effectiveness, recent studies~\citep{Liu2021WhatMG,Min2022RethinkingTR,an2023context,Lu2021FantasticallyOP,Zhou2023TheMO} have shown that ICL is sensitive to alterations in permutation, format, and quantity of provided demonstrations.

Addressing the order sensitivity of in-context examples and bridging the performance gap between optimal and other permutations presents a significant challenge, with discrepancies reaching up to 30\%~\citep{Lu2021FantasticallyOP}. Consequently, there are some studies aim to identify optimal permutations of in-context examples~\citep{Li2023FindingSE,Wu2022SelfAdaptiveIL,Scarlatos2023RetICLSR,Zhang2022ActiveES}. However, the majority of these studies focus solely on CausalLMs, where an auto-regressive attention mask is applied to limit token access to future token information in an input sequence. Moreover, they primarily focus on identifying the optimal order without fundamentally enhancing the robustness of LLMs towards variations in demonstration order. In this paper, we propose an unsupervised fine-tuning method to enhance the LLMs' robustness against different demonstration permutations. 

This method is inspired by the insights drawn from our preliminary investigation into both CausalLMs and PrefixLMs. 
PrefixLMs~\citep{Tay2022UL2UL,Raffel2019ExploringTL} employ full attention within the input tokens, allowing in-context samples all attend to each other.
For given demonstrations (i.e., in-context examples), we evaluate model performance across various permutations of the in-context examples, and introduce a metric termed as  `partial correct ratio'. 
This metric reflects the proportion of samples among those that can obtain the correct answer through majority voting, yet specific permutations lead to correct answers, whereas others result in incorrect ones.
We find that PrefixLMs exhibit significantly lower sensitivity to order compared to CausalLMs, with only 6.7\% of test cases showing partial correctness for Flan-T5-XXL, as opposed to 58.4\% for Llama2-chat-13B on the SST-5~\citep{Socher2013RecursiveDM} dataset. 

We attribute this phenomenon to the auto-regressive attention mask used in CausalLMs, which results in diverse receptive fields for a given sample depending on its position within the demonstration.
For example, suppose that the demonstration consists of 10 in-context examples and the test query is followed by the demonstration. 
When an in-context example is placed in the first position in the demonstration, the autoregressive attention mask prevents it from recognising the presence of the other 9 examples. 
Conversely, if it is placed in the last position, it can perceive others and interact with them via the self-attention mechanism.
Therefore, the positioning of an example within the demonstration significantly influences the information it can gather. 
Examples positioned further back in the demonstration possess richer information as they can perceive more examples, leading to an example's representation being greatly influenced by its position.
Despite the test query having access to the preceding 10 in-context examples, the representations of these in-context examples vary due to their different positions. This results in discrepancies in the attention calculation for the test query.

To alleviate the sensitivity of CausalLMs to the order of in-context examples, we propose a novel \emph{Information-Augmented and Consistency-Enhanced fine-tuning} approach~(\textbf{InfoAC})~\footnote{Our code and data are available at: \href{https://github.com/xyzCS/InfoAC}{https://github.com/xyzCS/InfoAC}}, which is simple, unsupervised and based on Low-Rank Adaptation~(LoRA)~\citep{Hu2021LoRALA}.
\textbf{InfoAC} consists of two components:
(1) \textbf{Information augmentation}: Recognizing that the vector representation of an in-context example is most informative when it is positioned last, and less so when it is positioned earlier, we propose using contrastive learning to align the representations of examples of earlier positions more closely with that of the last-positioned example. This adjustment ensures that regardless of their position, the examples encapsulate the full spectrum of information.
(2) \textbf{Consistency enhancement}: To promote consistency in predicting answers for a test sample with different permutations of in-context examples, we introduce a consistency loss. This loss function constrains the representations preceding the classification head to be similar across inputs corresponding to different permutations, enhancing the model's overall consistency in prediction.
During the inference stage, the proposed \textbf{InfoAC} exhibits strong generalizability, particularly when the in-context examples are chosen from a different pool than that utilized during training (Cross-Pool generalizability), or when the count of in-context examples differs from the training setup (Cross-Count generalizability).

In summary, our main contributions are:
\begin{itemize} 
	\item[$\bullet$] 
	 We discover that CausalLMs are more sensitive to the order of in-context examples compared to PrefixLMs based on our extensive experiments on SST-5 and MMLU datasets. We hypothesize that this discrepancy may be attributed to the auto-regressive attention mask employed within CausalLMs.
	
	\item[$\bullet$] To address this issue, we propose an unsupervised approach for fine-tuning, termed Information-Augmented and Consistency-Enhanced fine-tuning. This method aims to mitigate the impact of the order of in-context examples on the in-context learning performance of CausalLMs. 
	
	\item[$\bullet$] Experimental results across five benchmark datasets, employing a range of CausalLMs, reveal that the proposed \textbf{InfoAC} can effectively mitigate the order sensitivity of in-context examples. Besides, it showcases significant generalizability across different selected demonstration pools and the number of demonstrations.
\end{itemize}

%% file: sections/2_preliminary_experiments.tex
\begin{table*}[h]
\centering
\resizebox{0.7\linewidth}{!}{
\begin{tabular}{l|ccc|ccc}
	\toprule
	Benchmark & \multicolumn{3}{c|}{SST-5} & \multicolumn{3}{c}{MMLU} \\
	\midrule
	Metrics & MV$(\uparrow)$ & Partial$(\downarrow)$ & Entropy$(\downarrow)$ & MV$(\uparrow)$ & Partial$(\downarrow)$ & Entropy$(\downarrow)$ \\
	\midrule
	Flan-T5-XL & 0.489  & 0.191 & 0.127 & 0.478 & 0.100 & 0.077     \\
	Flan-T5-XXL & 0.364 & 0.067 & 0.046 & 0.530 & 0.111 & 0.084     \\
	Flan-UL2 & 0.550 &  0.101 & 0.062   & 0.538 & 0.098 & 0.075     \\
 	\midrule
	Llama-7B  & 0.490 & 0.557 & 0.399 & 0.476 & 0.237 & 0.195 \\
	Llama-13B & 0.483 & 0.584 & 0.437 & 0.532 & 0.195 & 0.167 \\
	Vicuna-7B & 0.494 & 0.449 & 0.315 & 0.484 & 0.247 & 0.216 \\
	Vicuna-13B & 0.502 & 0.541 & 0.386 &0.552 & 0.183 & 0.158 \\
	\bottomrule
\end{tabular}}
\caption{\label{PreExperimentResult10Shot}
	Experimental results on SST-5 and MMLU benchmarks. PrefixLMs are in the upper group, and CausalLMs are in the lower group.}
\end{table*}

\section{Order Sensitivity of PrefixLMs vs. CausalLMs}
\label{sec:pre_study}
\subsection{PrefixLMs and CausalLMs}
Transformer~\cite{Vaswani2017AttentionIA} has been widely adopted as the fundamental block of both the PrefixLMs and CausalLMs.  
Each Transformer block is composed of two components: a standard \textbf{S}oftmax \textbf{S}elf-\textbf{A}ttention layer~(SSA) and a \textbf{F}eed-\textbf{F}orward layer~(FF). Let $\bm{T} \in \mathbb{R}^{n \times d} = [\mathbf{t_1},\mathbf{t_2},...,\mathbf{t_n}]$ be the input, $n$ is the sequence length, and $d$ is the feature dimension. The SSA layer can be formulated as follows:

\begin{equation} \small
	\mathbf{t_i^{'}} \leftarrow \mathbf{t_i}+\text{Softmax}(\mathbf{t_i}W_QW_K\bm{T}^T+\textbf{Mask}_i)\bm{T}W_VW_O,
\end{equation}
\begin{equation} \small
	\label{MaskPrefix}
	\textbf{Mask}_i^{\text{PrefixLM}} = \bm{0}^{1 \times n}
\end{equation}
\begin{equation} \small
	\label{CausalPrefix}
	\textbf{Mask}_i^{\text{CausalLM}} =
	\begin{cases}
	\textbf{Mask}_i^{\text{CausalLM}}[j]=0,  & \text{if $j \leq i$} \\
	\textbf{Mask}_i^{\text{CausalLM}}[j]=-\infty, & \text{if  $j > i$}
	\end{cases}
\end{equation}
where $W_Q, W_K, W_V, W_O\in \mathbb{R}^{d 
\times d}$ are trainable parameters corresponding to the query, key, value, and output projections, respectively. 
$\textbf{Mask}_i \in \mathbb{R}^{1 
\times n}$ refers to the attention mask. 
The main difference between CausalLMs and PrefixLMs lies in the mask embedding.
The mask embedding $\textbf{Mask}_i^{\text{PrefixLM}}$ for PrefixLMs is represented as a zero matrix, as indicated in equation~\ref{MaskPrefix}.
In a CausalLM, the auto-regressive mask embedding $\textbf{Mask}_i^{\text{CausalLM}}$ for the i-th token is set to 0 for positions before or at the i-th token and negative infinity for positions following it, as specified in equation~\ref{CausalPrefix}.
After passing through the softmax activation function, the weights for tokens after the i-th token are zero, thus preventing the i-th token from accessing information from subsequent tokens.
The FF layer is followed by the SSA layer: 
\begin{equation}
	 \mathbf{t_i^{''}} \leftarrow \text{GELU}(\mathbf{t_i^{'}}W_1)W_2,
\end{equation}
where $W_1$ and $W_2$ are trainable parameters and $\text{GELU}(.)$ denotes the GELU~\citep{Hendrycks2016GaussianEL} activation function.

\subsection{Preliminary Experimental Results}
\label{sec:pre_study}
In this sub-section, we investigate the sensitivity of PrefixLMs and CausalLMs to the order of in-context demonstration examples. 
The details of the chosen models are shown in Table~\ref{ModelInfo}.
We conduct experiments on the \textbf{S}tanford \textbf{S}entiment \textbf{T}reebank with 5 labels (SST-5)~\citep{Socher2013RecursiveDM} and \textbf{M}assive \textbf{M}ultitask \textbf{L}anguage \textbf{U}nderstanding (MMLU)~\citep{Hendrycks2020MeasuringMM} benchmarks.
MMLU benchmark encompasses 57 subtasks, covering a wide range of areas including STEM, the humanities, and beyond.  
We exclude sub-tasks that would surpass Flan-T5's~\citep{Raffel2019ExploringTL} maximum input length and present the average metrics for the remaining 43 datasets.
For each test case, we randomly select 10 samples from the validation set as its in-context examples.
Given that there are 10 factorial possible permutations for 10 in-context samples, testing every permutation is impractical. 
Therefore, we sample 20 permutations for each test case to create prompts and fed them to LLMs to obtain their predicted answers.

\paragraph{Metrics}
\label{metrics}
To measure the model's sensitivity to the ordering of in-context examples, we introduce three metrics: Majority Voting Accuracy, Partial Correct Ratio, and Entropy.
\begin{itemize}  [leftmargin=0pt,itemsep=-3pt,topsep=2pt]   
	\item \textsl{Majority Voting Accuracy (MV)}: It employs majority voting to consolidate the results of all prompts for a test case to obtain the final answer, and then computes the proportion of the test cases where the correct answer can be found. 
    \item \textsl{Partial Correct Ratio (Partial)}: It represents the proportion of samples, among those that can obtain the correct answer through majority voting, for which only certain permutations, and not all permutations, can yield the correct answer.
    \item \textsl{Entropy}: For each test case, a count of distinct answers is conducted across all prompts. These counts are normalised by dividing by the total number of prompts to obtain a discrete distribution over the label space. The entropy of this distribution is then calculated to measure the level of disorder in the prediction results. Higher entropy implies less uniformity in the predicted answers. We report the average entropy of test cases that could obtain the correct answer through majority voting.
\end{itemize}

\paragraph{Results}
Table~\ref{PreExperimentResult10Shot} reports the 10-shot results for the SST-5 and MMLU dataset.
We also report results when the number of in-context examples is set to 4 and apply all 24 permutations to construct prompts, as detailed in Table~\ref{PreExperimentResult4Shot}.
It is evident that CausalLMs exhibit a higher sensitivity to the order of in-context examples compared to PrefixLMs.
As indicated in Table~\ref{PreExperimentResult10Shot}, for the MMLU dataset, the partial correct ratio metric for all PrefixLMs remains around 0.1, whereas for the four CausalLMs, this metric can exceed 0.2, which is twice that of the PrefixLMs.
Additionally, the entropy metric for PrefixLMs stays below 0.1, while for Llama-7B and Vicuna-7B, the entropy values are significantly higher, recorded at 0.195 and 0.216, respectively.
This discrepancy suggests that CausalLMs are more inclined to produce varied predictions when confronted with different permutations of in-context examples.

\begin{figure}[t]
 	\centering
 	\includegraphics[width=0.40\textwidth]{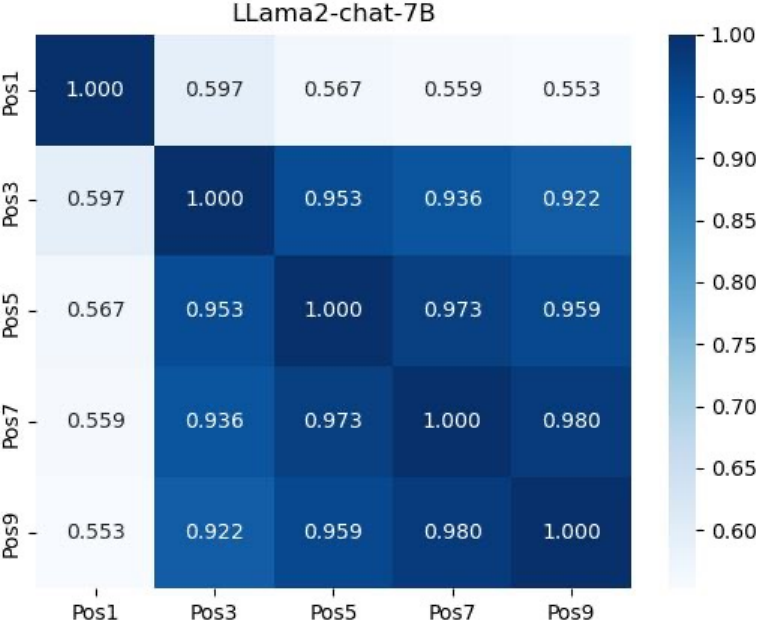} 
 	\caption{\label{heatLLama}
 	The heatmap visualizes the similarities in representations of a specific token within a sample from the last layer outputs across different positions for Llama2-chat-7B.
 	}
 \end{figure}
 
 \begin{figure}[t]
 	\centering
 	\includegraphics[width=0.40\textwidth]{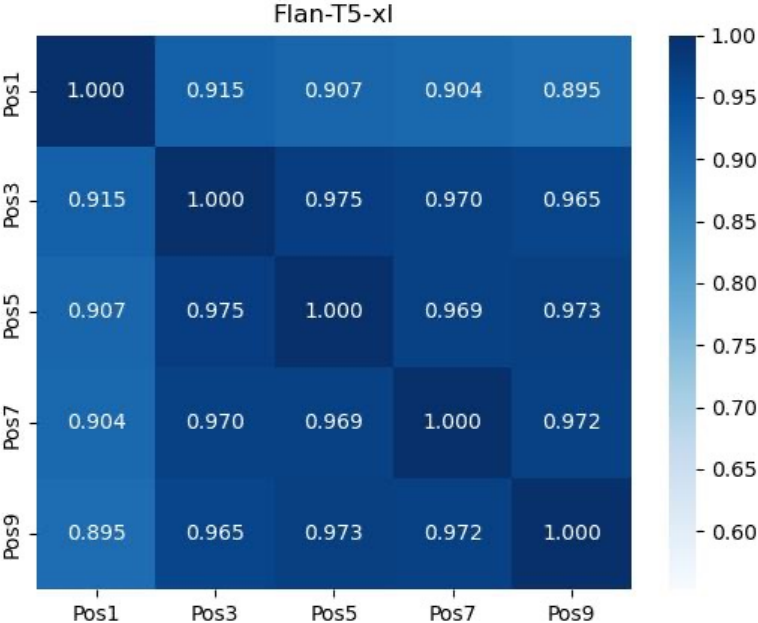} 
 	\caption{\label{heatT5}
 	The heat map visualises the similarities in representations  of a specific token within a sample from the last encoder layer outputs across different positions for Flan-T5-XL.}
 \end{figure}
 
\paragraph{Analysis}
To delve deeper into the order sensitivities of PrefixLMs and CausalLMs from representation perspectives, we compare the representations of a token within a particular in-context sample when it is placed in different positions.
Figure~\ref{heatLLama} and Figure~\ref{heatT5} respectively show the heatmaps of the similarities in representations for the Llama2-chat-7B and Flan-T5-XL models, when a sample is placed in different positions.
Specifically, we sample 10 in-context examples and organise them in a random permutation.
For the sample placed in the first position, we gradually move it backwards while keeping the relative positions of the other samples unchanged.
Next, we retrieve the representations of a specific token within the sample from the LLMs at each position, utilizing the output from the final encoder layer of Flan-T5 and the output from the last layer of Llama.
We calculate pairwise similarities for representations placed in odd positions.
This process is repeated 100 times to ensure robustness. 
The average of these similarity matrices is then calculated, and the resulting heatmap is plotted based on this average similarity matrix.
Clearly, the similarity of the token across various positions is less for Llama than for Flan-T5-XL.
For Flan-T5-XL, the similarity across representations at any position typically exceeds 0.90, while for Llama, the lowest value is approximately 0.553. 
This observation aligns with our evaluation results in Table~\ref{PreExperimentResult10Shot}, indicating that CausalLMs are more sensitive to position variations.
In particular, when the sample is positioned at the first position, the representation of the token shows the greatest difference compared to later positions, as it is unable to ``see'' other samples in the CausalLMs due to their position-dependent visibility.

Based on the observations, we propose \textbf{InfoAC} to minimise the discrepancy of representations across different positions.
As illustrated in Figure~\ref{heatLLamaInfoAC}, the application of our proposed \textbf{InfoAC} to CausalLMs has led to consistent representations of the sample across various positions.

%% file: sections/3_method.tex
\section{Method}

\begin{figure}[t]
	\centering
	\includegraphics[scale=0.55]{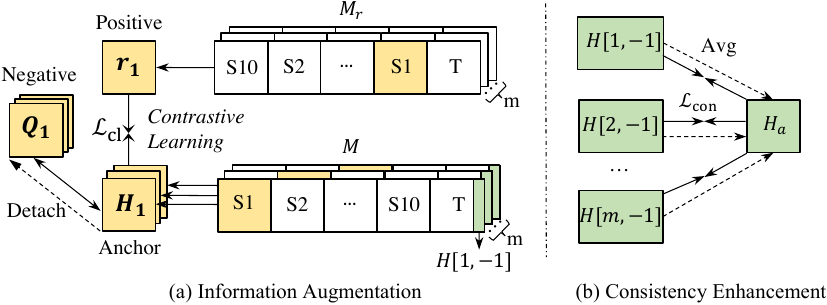} 
	\caption{\label{model}
        The overview of our proposed \textbf{InfoAC}. We adopt contrastive learning~(Left) to align the representation of a sample, $S_1$, as derived from model $M$, with the representations of $S_1$ when it is positioned at the end of the sequence derived from a referenced model $M_r$. We also ensure that the hidden representations preceding the classification head are similar when positioned at various locations, resulting in consistent outputs~(Right).
	}
\end{figure}

\subsection{Task Definition}
In-context learning involves learning a conditional text generation probability given a demonstration and a test query.
First, a subset of the training set is selected and reserved as a candidate pool $\mathbb{P}=\{(x_i, y_i)\}_{i=1}^{N_P}$, for selecting in-context examples~\citep{Chang2022DataCA,Nguyen2023IncontextES,Li2023FindingSE,Wu2022SelfAdaptiveIL}.
Then, given a test query $q$, a LLM $M$ generates a probability distribution for target $y$ conditioned on the demonstration $C$:
\begin{equation}
	P_{M} =  \prod \limits_{t=0}^{n_c} p(y_t|y_{<t},q,C),
\end{equation}
where $n_c$ is the length of the generated sequence, $y_{t}$ is an output label token at each time step.
The demonstration $C$ consists of $k$ input-label pairs $\mathbb{D} =\{(x_1, y_1),...,(x_k,y_k)\}$ chosen from $\mathbb{P}$ and organised according to a specific permutation $f_i \in \mathbb{F}$.
Here, $\mathbb{F}=\{f_i\}$, $i=1,...,k!$ denotes all possible permutations.

\subsection{Information-Augmented and Consistency-Enhanced Fine-Tuning}
\label{sec:infoAC}
As previously discussed, the notable order sensitivity of in-context examples observed in CausalLMs is likely attributed to the auto-regressive attention masks, which results in discrepancies in their contextual representations based on their positions.
To deal with this problem, a direct approach is to modify the attention mask of CausalLMs to resemble that of PrefixLMs' attention mask.
However, modifying the mask embeddings requires significant changes to the model architecture, demanding a considerable amount of data for training~\cite{Song2020MPNetMA}.
Attempting to implement this change during the inference stage without training would lead to significant performance degradation.

Therefore, we propose to instead mitigate the discrepancy in representations caused by different positions through unsupervised training objectives. The method is motivated by two characteristics of position-invariant representations: (1) the representation of a given in-context example should remain robust regardless of the permutations of other input examples, and (2) it should be aware of the information contained within all other examples. 
Guided by these two principles, our training objective aims to align representations of an in-context example across different positions with that at the end of the demonstration. 
This end-point representation encapsulates information from preceding demonstrations in the CausalLMs, hence termed as \textsl{Information Augmentation}. 
Furthermore, we enhance representation consistency near the prediction head by ensuring similarity among representations stemming from different demonstration permutations, referred to as \textsl{Consistency enhancement}. 
The overview of the proposed approach is shown in Figure~\ref{model}.
We adopted the LoRA adaptation\citep{Hu2021LoRALA}, where all model parameters are fixed, except for the trainable query $W_Q$ and value $W_V$ matrices within the self-attention layers. In the following subsections, we will introduce the unlabelled training data and the two key components.

\subsubsection{Training and Test Data Construction}
We first randomly chose $N_P$ samples from the training set to create the candidate pool $\mathbb{P}$.
Then, from the remaining training set, $N_T$ samples are selected to form a pseudo test set, denoted as $\mathbb{P}_T$.
In order to construct a training batch $I=[I_1,I_2,...,I_m]$, we sample $k$ in-context examples randomly from $\mathbb{P}$, a pseudo test sample from $\mathbb{P}_T$ and $m$ different permutations from $\mathbb{F}$.
Subsequently, $m$ prompts are generated by organizing each prompt according to the corresponding permutations obtained.
Since the representation of the sample in the first position has the greatest difference from the representations in other positions, we reverse the permutation of the sampled permutations and obtain the input batch $I_r$ for the reference model $M_r$. 
This way, every sample placed in the first position in $I$ has a reference that is placed at last in $I_r$.
To maximize the diversity of references for different samples, we ensure that the first in-context example in each input of a training batch $I$ is unique.
We repeat the sampling process $N$ times to obtain our training data and reference data for the training model $M$ and the reference model $M_r$.
As for the test set, we randomly select $k$ in-context examples from $\mathbb{P}$ for each test case and sample 20 permutations to construct prompts. 
After enumerating all test cases, we obtain the test set.

\subsubsection{Information Augmentation}
Recall that our objective is to align the representations from different positions with those at the end of the demonstration. 
To achieve this, we employ an original version of the model as a reference model $M_r$ to derive the reference representation.
Notably, $M_r$ is fixed and doesn't require gradient update during the fine-tuning process.
By feeding $I$ and $I_r$ into $M$ and $M_r$, we obtain the token-level outputs of the last self-attention layer, denoted as $\bm{H} \in \mathbb{R}^{m \times n \times d}$ and $\bm{H_r} \in \mathbb{R}^{m \times n \times d}$, respectively:
\begin{equation}
	\bm{H} = M(I), \bm{H_r} = M_r(I_r),
\end{equation}
Here, $m$ is the number of sampled permutations and also corresponds to the batch size, $n$ indicates the length of the input sequence, and $d$ refers to the dimension of the output representation.
Then we obtain reference representations $\mathbb{R}= \{\bm{r_1}, \bm{r_2},...,\bm{r_m}\}$ for those in-context examples $\mathbb{S} = \{s_1, s_2,..., s_m\}$ that are placed at the last position of the demonstration according to Eq~(\ref{eq:avg_reference}):
\begin{equation}
	\bm{r_i} = \bm{H_r}[i,\text{Start}_{i}^{i}:\text{End}_{i}^{i}],
\label{eq:avg_reference}
\end{equation}
where $\text{Start}_{i}^{i}$ and $\text{End}_{i}^{i}$ denote the start and the end index of $s_i$ within the input $I_i$.
We introduce a token-level contrastive learning objective $\mathcal{L}_{cl}$ to integrate the information from the reference representation into the learning process of the model $M$:
\begin{equation} \small
\mathcal{L}_{cl} =  -\text{log}\sum_{i=1}^{m}\sum_{j}^{s_j \in \mathbb{S}} 
	 \sum_{o=1}^{n_j} \frac{\text{Pos}(i,j,o)}{\text{Pos}(i,j,o) + \text{Neg}(i,j,o)},
\end{equation}
\begin{equation} \small
\text{Pos}(i,j,o) = \text{exp}\big(\text{cos}(\bm{H}[i,\text{Start}_{i}^{j}+o],
					  \bm{r}_j[i,o])\big)/\tau, \\
\end{equation}

\begin{equation} \small
\text{Neg}(i,j,o) =  \text{exp}\big(\text{cos}(\bm{H}[i,\text{Start}_{i}^{j}+o], 
					   \bm{Q})\big)/\tau.
\end{equation}
Here, $n_j$ is the number of tokens within the in-context example $s_j$, $\text{Start}_{i}^{j}$ represents the start index of $s_j$ within the input $I_i$ and $\text{cos}(.)$ denotes the cosine similarity function.
The positive sample $\bm{r}_j[i,o]$ is the reference representation of the token, while the negative sample $\bm{Q}$ is an independent copy of $\bm{H}[i,\text{Start}_{i}^{j}+o]$, detached from the computational graph.

This loss mechanism aims to align the self-attention output of each token more closely with the reference output, thereby increasing the distance from its original representation.
With this approach, the model integrates information from the reference representation into the LoRA parameters of self-attention layers while fine-tuning.
Consequently, it allows a token to access information from subsequent tokens, regardless of the limitations imposed by the attention masks.

\subsubsection{Consistency Enhancement}
For all inputs $I$ within a batch, we aim for the model's predicted results to exhibit consistency.
As the LoRA adaptation is applied, the parameters of the classification head remain fixed during training.
Thus, we enforce similarity among the hidden representations of the last token across the input batch $I$, which are then input into the classification head to predict the next output tokens.
This approach is designed to enhance the consistency of the model's output.
\begin{equation}
\mathcal{L}_{\text{con}}=\sum_{i=1}^{m} (1-\text{cos}(\bm{H}[i,-1], \bm{H_a})),
\end{equation}
\begin{equation}
\bm{H_a}= \frac{\sum_{i=1}^{m}{\bm{H}[i,-1]}}{m}, 
\end{equation}
Here, $\bm{H_a} \in \mathbb{R}^{d}$ represents the average representation derived from all inputs. 
The total loss $\mathcal{L}$ combines these two losses:
\begin{equation}
\mathcal{L}= \mathcal{L}_{cl} + \lambda_1 \mathcal{L}_{\text{con}}
\end{equation}

%% file: sections/4_experiment.tex
\begin{table*}[t]
\centering
\resizebox{0.8\linewidth}{!}{
\begin{tabular}{l|ccc|ccc}
	\toprule
	Dataset	& \multicolumn{3}{c|}{SST-5}    & \multicolumn{3}{c}{QQP}          \\
	\midrule
	Metrics & MV($\uparrow$)  & Partial($\downarrow$) & Entropy($\downarrow$) & MV($\uparrow$)  & Partial($\downarrow$) & Entropy($\downarrow$)  \\
	\midrule
	Llama-7B  & 0.501$\pm$0.009 & 0.566$\pm$0.013 & 0.404$\pm$0.017 & 0.665$\pm$0.002 & 0.749$\pm$0.022 & 0.503$\pm$0.008 \\
	+IA & 0.512$\pm$0.009 & 0.271$\pm$0.021 & 0.179$\pm$0.013 & \textbf{0.702}$\pm$\textbf{0.002} & 0.371$\pm$0.065 & 0.245$\pm$0.042 \\
	+InfoAC & \textbf{0.518}$\pm$\textbf{0.013} & \textbf{0.159}$\pm$\textbf{0.004} & \textbf{0.105}$\pm$\textbf{0.004} & 0.697$\pm$0.016 & \textbf{0.095}$\pm$\textbf{0.066} & \textbf{0.062}$\pm$\textbf{0.046}\\
	\midrule
	Llama-13B  & 0.491$\pm$0.007 & 0.605$\pm$0.024 & 0.454$\pm$0.019 & 0.720$\pm$0.019 & 0.501$\pm$0.028 & 0.300$\pm$0.027 \\
	+IA & 0.497$\pm$0.006 & 0.297$\pm$0.022 & 0.197$\pm$0.014 & \textbf{0.735}$\pm$\textbf{0.024} & 0.255$\pm$0.062 & 0.162$\pm$0.036\\
	+InfoAC & \textbf{0.505}$\pm$\textbf{0.004} & \textbf{0.171}$\pm$\textbf{0.011} & \textbf{0.110}$\pm$\textbf{0.007} & 0.718$\pm$0.015 & \textbf{0.117}$\pm$\textbf{0.030} & \textbf{0.077}$\pm$\textbf{0.021}\\
	\midrule
	Vicuna-7B  & 0.502$\pm$0.006 & 0.469$\pm$0.029 & 0.330$\pm$0.027 & 0.681$\pm$0.003 & 0.681$\pm$0.034 & 0.437$\pm$0.033 \\
	+IA & \textbf{0.503}$\pm$\textbf{0.003} & 0.210$\pm$0.020 & 0.138$\pm$0.017 & \textbf{0.684}$\pm$\textbf{0.016} & 0.380$\pm$0.088 & 0.252$\pm$0.059 \\
	+InfoAC & 0.500$\pm$0.005 & \textbf{0.123}$\pm$\textbf{0.015} & \textbf{0.081}$\pm$\textbf{0.014} & 0.678$\pm$0.005 & \textbf{0.048}$\pm$\textbf{0.044} & \textbf{0.030}$\pm$\textbf{0.030} \\
	\midrule
	Vicuna-13B  & 0.493$\pm$0.009 & 0.524$\pm$0.024 & 0.381$\pm$0.023 & 0.746$\pm$0.009 & 0.452$\pm$0.033 & 0.284$\pm$0.026 \\
	+IA & 0.505$\pm$0.005 & 0.320$\pm$0.029 & 0.219$\pm$0.027 & \textbf{0.770}$\pm$\textbf{0.016} & 0.314$\pm$0.074 & 0.203$\pm$0.045 \\
	+InfoAC & \textbf{0.511}$\pm$\textbf{0.008} & \textbf{0.223}$\pm$\textbf{0.046} & \textbf{0.149}$\pm$\textbf{0.033} & 0.740$\pm$0.0205 & \textbf{0.205}$\pm$\textbf{0.045} & \textbf{0.131}$\pm$\textbf{0.028}\\
	\midrule
	Dataset	& \multicolumn{3}{c|}{Sequence Next Term}   & \multicolumn{3}{c}{Round Number} \\ 
	\midrule
	Metrics & MV($\uparrow$) & Partial($\downarrow$) & Entropy($\downarrow$) & MV($\uparrow$) & Partial($\downarrow$) & Entropy($\downarrow$)  \\
	\midrule
	Llama-7B & 0.246$\pm$0.020 & 0.379$\pm$0.008 & 0.352$\pm$0.019 & 0.205$\pm$0.012 & 0.353$\pm$0.015 & 0.251$\pm$0.007\\
	+IA & 0.255$\pm$0.008 & 0.199$\pm$0.020 & 0.149$\pm$0.029 & 0.208$\pm$0.004 & 0.139$\pm$0.015 & 0.090$\pm$0.010\\
	+InfoAC & \textbf{0.257}$\pm$\textbf{0.005} & \textbf{0.174}$\pm$\textbf{0.029} & \textbf{0.136}$\pm$\textbf{0.025} & \textbf{0.208}$\pm$\textbf{0.006} & \textbf{0.099}$\pm$\textbf{0.017} & \textbf{0.070}$\pm$\textbf{0.016}\\
	\midrule
	Llama-13B & 0.260$\pm$0.026 & 0.348$\pm$0.056 & 0.306$\pm$0.058 & 0.244$\pm$0.015 & 0.412$\pm$0.067 & 0.311$\pm$0.056\\
	+IA & 0.281$\pm$0.007 & 0.168$\pm$0.014 & 0.128$\pm$0.008 & 0.248$\pm$0.002 & 0.164$\pm$0.032 &  0.109$\pm$0.020\\
	+InfoAC & \textbf{0.281}$\pm$\textbf{0.009} & \textbf{0.134}$\pm$\textbf{0.027} & \textbf{0.107}$\pm$\textbf{0.011} & \textbf{0.249}$\pm$\textbf{0.005} & \textbf{0.130}$\pm$\textbf{0.020} & \textbf{0.091}$\pm$\textbf{0.008}\\
	\midrule
	Vicuna-7B & 0.226$\pm$0.002 & 0.499$\pm$0.018 & 0.579$\pm$0.046 & \textbf{0.199}$\pm$\textbf{0.007} & 0.391$\pm$0.041 & 0.302$\pm$0.034\\
	+IA & 0.226$\pm$0.010 & 0.257$\pm$0.061 & 0.234$\pm$0.064 & 0.193$\pm$0.002 & 0.181$\pm$0.036 & 0.118$\pm$0.017\\
	+InfoAC & \textbf{0.227}$\pm$\textbf{0.008} & \textbf{0.239}$\pm$\textbf{0.040} & \textbf{0.208}$\pm$\textbf{0.031} & 0.192$\pm$0.007 & \textbf{0.160}$\pm$\textbf{0.029} & \textbf{0.107}$\pm$\textbf{0.018}\\
	\midrule
	Vicuna-13B & 0.263$\pm$0.026 & 0.380$\pm$0.056 & 0.360$\pm$0.062 & 0.250$\pm$0.010 & 0.393$\pm$0.050 & 0.287$\pm$0.061\\
	+IA & 0.277$\pm$0.002 & 0.220$\pm$0.013 & 0.183$\pm$0.029  & 0.249$\pm$0.012 & 0.266$\pm$0.030 & 0.180$\pm$0.020\\
	+InfoAC & \textbf{0.278}$\pm$\textbf{0.001} & \textbf{0.189}$\pm$\textbf{0.025} & \textbf{0.158}$\pm$\textbf{0.030} & \textbf{0.250}$\pm$\textbf{0.011} & \textbf{0.210}$\pm$\textbf{0.024} & \textbf{0.149}$\pm$\textbf{0.015}\\
	\bottomrule
\end{tabular}
}
\caption{\label{MainResults1}
	Experimental results on SST-5, SST-2, Sequence Next Term and Round Number benchmarks.
	``\textbf{IA}'' denotes the ``Information augmentation'', while ``\textbf{InfoAC}'' further introduces the consistency enhancement.}
\end{table*}

\section{Experiment}
\subsection{Experimental Setup}
We conduct our experiments using 5 datasets that cover a range of types such as text classification, textual entailment, and mathematics.
These datasets include SST-5~\citep{Socher2013RecursiveDM}, SST-2~\citep{Socher2013RecursiveDM}, QQP~\citep{quora-question-pairs}, Sequence Next Term~\citep{2019arXiv} and Round Number~\citep{2019arXiv}. 
The statistics for each dataset are presented in Table~\ref{Datasets}, and detailed descriptions of each dataset can be found in section~\ref{Benchmark}.

The details of the experimental setup for training, inference, and in-context learning are provided in section~\ref{ExperimentalSetup}. 
 We use the same evaluation metrics, i.e., Majority Voting Accuracy (MV), Partial Correct Ratio (Partial) and Entropy, as outlined in the section~\ref{sec:pre_study}. 

\subsection{Main Results}
To mitigate the effects of different candidate pools, we perform the experiment three times, randomly selecting a new candidate pool for each iteration.
Table~\ref{MainResults1} reports the mean and standard deviation of the corresponding metrics across three iterations on SST-5, QQP, Sequence Next Term and Round Number benchmarks while the results on the SST-2 benchmark are reported in the Table~\ref{MainResults2}.

Our observations are as follows:
(1) Across the five datasets, all four backbone models clearly show sensitivity to the ordering of in-context examples, indicating that order sensitivity is independent of the task type and domain. 
Specifically, for the SST-5 and QQP datasets, the partial correct ratio metrics for all models are around or exceed 0.5. 
(2) As the model size increases, it does not reduce sensitivity to order. 
For instance, in the SST-5 dataset, Llama-13B and Vicuna-13B have partial correct ratios of 0.605 and 0.524, respectively. 
These metrics are higher compared to 0.566 for Llama-7B and 0.469 for Vicuna-7B.
A similar trend is also observed in the Round Number benchmark.
(3) Our proposed Information Augmentation (\textbf{IA}) method significantly mitigates the sensitivity of CausalLMs to order, with an average decrease of about 0.192 and 0.157 on the partial correct ratio and the entropy metric. 
Even though it is unsupervised, \textbf{IA} can enhance overall performance to some degree on certain datasets. 
For instance, it results in an average increase of 0.020 in the majority voting accuracy metric on the QQP dataset.
(4) Introducing the consistency loss further enhances the robustness of CausalLMs to the order of in-context examples. 
Compared to the \textbf{IA} method, it results in a 0.079 decrease in the partial correct ratio and a 0.054 decrease in the entropy metrics.

\begin{table*}[t]
\centering
\resizebox{0.7\linewidth}{!}{
\begin{tabular}{l|ccc|ccc}
	\toprule
	Metrics & MV$(\uparrow)$  & Partial$(\downarrow)$ & Entropy$(\downarrow)$ & MV$(\uparrow)$  & Partial$(\downarrow)$ & Entropy$(\downarrow)$ \\
	\midrule
	& \multicolumn{3}{c|}{KNN} & \multicolumn{3}{c}{One-Shot} \\ 
	\midrule
	Llama-7B & 0.514 & 0.581 & 0.409 & 0.501 & 0.594 & 0.432  \\
	+\textbf{InfoAC}  & \textbf{0.521} &  \textbf{0.162} &  \textbf{0.111}  & \textbf{0.511} & \textbf{0.159} & \textbf{0.100}	  \\
	\midrule
	Llama-13B & 0.503 & 0.623 & 0.440 & 0.496 & 0.597 & 0.448  \\
	+\textbf{InfoAC} & \textbf{0.520} & \textbf{0.188} & \textbf{0.124} & \textbf{0.507} & \textbf{0.194} & \textbf{0.129}    \\
	\midrule
	Vicuna-7B &  \textbf{0.503} & 0.445 & 0.318 & 0.515 & 0.516 & 0.374 \\
	+\textbf{InfoAC} & 0.495 &  \textbf{0.142} &  \textbf{0.097} & \textbf{0.518} & \textbf{0.199} & \textbf{0.135}\\
	\midrule
	Vicuna-13B & 0.518 & 0.546 & 0.383 & 0.511  & 0.390 & 0.263  \\
	+\textbf{InfoAC} &  \textbf{0.531} &  \textbf{0.234} &  \textbf{0.153} &  \textbf{0.518} & \textbf{0.106} & \textbf{0.071} \\
	\bottomrule
\end{tabular}}
\caption{\label{InContextSelectionSST5}
	Performances of different in-context sample selection methods on the SST-5 dataset.
	}
\end{table*}

\subsection{In-context Selection Results}
In addition to random in-context example selection results presented above, we explore whether the advanced in-context example selection methods can enhance the robustness of CausalLMs to the order of in-context examples. 
Specifically, we focus on KNN and One-Shot approaches:
\begin{itemize} [leftmargin=0pt,itemsep=-3pt,topsep=2pt]
    \item \textbf{K-Nearest Neighbors~(KNN)}~\citep{Liu2021WhatMG,Gao2021MakingPL}: KNN aims to select the semantically closest samples from the candidate pool as in-context examples for each test case.  Given a test sample and a candidate pool, an LM is adopted to encode the test sample and each candidate sample. Next, the cosine similarity is computed between the test query's representation and the representations of all candidate queries. The samples with the highest similarity serve as in-context examples.
    \item \textbf{One-Shot}~\citep{Chang2022DataCA}: One-shot aims to find a candidate pool that contains higher-quality samples. It reserves a small set $D_{dev}$ from the training set to serve as a pseudo validation set. The remaining training samples are individually utilized as demonstrations for one-shot in-context learning, with each sample's score determined by its ICL accuracy on $D_{dev}$. The top-ranked samples are selected as the candidate pool. 
\end{itemize}
The implementation details are provided in Appendix~\ref{InContextSelectionDetails}.
Table~\ref{InContextSelectionSST5} presents the results of applying the KNN and One-shot approaches on the SST-5 dataset, while Table~\ref{InContextSelectionNext} displays the results on the Sequence Next Term dataset.
It's evident that employing more effective methods for selecting in-context examples can enhance the model's overall performance, as seen in the increase in majority voting accuracy. 
However, these improvements do not address the model's inherent sensitivity to the order of demonstrations.
The high sensitivity of CausalLMs to the order of demonstrations stems from the model's architecture, specifically the auto-regressive attention masks. 
After applying \textbf{InfoAC}, significant improvements are noticeable across a range of metrics, including majority voting accuracy.

\begin{table*}[t]
\centering
\resizebox{\linewidth}{!}{
\begin{tabular}{l|ccc|ccc|ccc}
    \toprule
    Metrics & MV$(\uparrow)$ & 
    Partial$(\downarrow)$ & Entropy$(\downarrow)$ & MV$(\uparrow)$ & 
    Partial$(\downarrow)$ & Entropy$(\downarrow)$ & MV$(\uparrow)$ & 
    Partial$(\downarrow)$ & Entropy$(\downarrow)$ \\
    \midrule
    & \multicolumn{3}{c|}{k=4} & \multicolumn{3}{c|}{k=20} & \multicolumn{3}{c}{k=50} \\
    \midrule
    Llama-7B & 0.504 & 0.556 & 0.390 & 0.501 & 0.592 & 0.428 & 0.506 & 0.643 & 0.446 \\
    +\textbf{InfoAC}  & \textbf{0.506} & \textbf{0.155} & \textbf{0.097} & 	 \textbf{0.515} & \textbf{0.224} & \textbf{0.153} & \textbf{0.519} & \textbf{0.385} & \textbf{0.245}   \\
    \midrule
    Llama-13B & 0.459 & 0.623 & 0.482 & 0.486 & 0.557 & 0.411 & 0.504 & 0.592 & 0.412  \\
    +\textbf{InfoAC} & \textbf{0.506} & \textbf{0.155} & \textbf{0.099} & \textbf{0.519} & \textbf{0.232} & \textbf{0.155} & \textbf{0.514} & \textbf{0.309} & \textbf{0.202}   \\
    \midrule
    Vicuna-7B & 0.496 & 0.353 & 0.262 & 0.500 & 0.466 & 0.331 & \textbf{0.514} & 0.521 & 0.367 \\
    +\textbf{InfoAC} & \textbf{0.500} & \textbf{0.110} & \textbf{0.078} & \textbf{0.505} & \textbf{0.200} & \textbf{0.137} & 0.507 & \textbf{0.266} & \textbf{0.176} \\
    \midrule
    Vicuna-13B & 0.477 & 0.516 & 0.399 & 0.509 & 0.558 & 0.391 & 0.490 & 0.547 & 0.399 \\
    +\textbf{InfoAC} & \textbf{0.513} & \textbf{0.225} & \textbf{0.156} & \textbf{0.522} & \textbf{0.253} & \textbf{0.162} & \textbf{0.516} & \textbf{0.374} & \textbf{0.246} \\
    \bottomrule
\end{tabular}}
\caption{\label{CrossLengthSST5}
	Cross-count experimental results on the SST-5 dataset. ``\textbf{InfoAC}'' represents the results employed with the proposed \textbf{InfoAC}, where the number of in-context examples $k$ is set to 10 during the fine-tuning phase.
 }
\end{table*}

\subsection{Generalizability Across Different Numbers of In-Context Examples and Different Candidate Pools}
In this section, we explore the generalizability of the CausalLMs after fine-tuning with the proposed \textbf{InfoAC}.
We first explore its generalizability across varying numbers of in-context samples.
In the previous experiments, we set the number of in-context examples $k$ to 10. 
Here, we directly apply the model trained with \textbf{InfoAC} at $k=10$ to testing at $k=4$, $k=20$ and $k=50$. 
The candidate pool $\mathbb{P}$ remains the same across the different values of $k$.
Table~\ref{CrossLengthSST5} and Table~\ref{CrossLengthNext} show the experimental results on the SST-5 and Sequence Next Term benchmarks.
We can observe that:
(1) As k increases, the model maintains its sensitivity to order and even shows a trend of increase. When $k$ increases from 4 to 20 and 50,  the partial correct ratio of Llama-7B on the SST-5 dataset increases from 0.556 to 0.592 and 0.643, respectively.
(2) The proposed \textbf{InfoAC} can generalize across different numbers of in-context samples. 
For example, when $k=20$, for LLama-7B, LLama-13B, Vicuna-7B and Vicuna-13B, after applying \textbf{InfoAC}, the partial correct ratio metric decreased by 0.368, 0.325, 0.266 and 0.305 respectively. 
Meanwhile, the entropy decreased by 0.275, 0.256, 0.194 and 0.229 respectively. 
A similar trend can also be observed when $k=4$ and $k=50$.
(3) When $k=4$, InfoAC not only reduces the model's sensitivity to order but also enhances the overall performance of the model. 
On the SST-5 and Sequence Next Term datasets, all four models show improvements in the majority voting metric. 
This is because the model is trained at $k=10$, using the representation at the 10th position to augment the representations at earlier positions, thereby improving performance.

We further explore the effectiveness of our method across different candidate pools in Appendix~\ref{CrossPool} which demonstrates that the CausalLMs fine-tuned with \textbf{InfoAC} have robust generalizability to different candidate pools.
It means when a different candidate pool is used during testing, the model still significantly reduces its sensitivity to the order of in-context examples. 

%% file: sections/5_related_work.tex
\section{Related Work}
Recently, some research has focused on investigating how the order of in-context examples affects in-context learning. 
\citet{Lu2021FantasticallyOP} proposed that the order of the demonstrations has a significant impact on the performance of in-context learning and the optimal order is model-dependent and not transferable across different models.
\citet{Liu2023LostIT} proposed that ICL achieves optimal performance when the relevant information is positioned at the beginning or end of the demonstrations.
The performance notably degrades when models are required to retrieve relevant information from the middle of long contexts. 
\citet{Bertsch2024InContextLW, Agarwal2024ManyShotIL} explore the order sensitivity of in-context examples in the scenario of many-shot in-context learning, where the demonstrations contain hundreds or even thousands of in-context examples.
\citet{Agarwal2024ManyShotIL} observes that LLMs still exhibit significant order sensitivity of in-context examples and \citet{Bertsch2024InContextLW} proposes that as the number of in-context examples increases, the order sensitivity decreases.

Therefore, numerous research efforts focus on finding the best permutations of in-context examples.
\citet{Wu2022SelfAdaptiveIL} proposed a ranking algorithm based on the perspective of compression. 
The assumption is that a well-organized arrangement of in-context examples can effectively compress test samples without sacrificing any information.
\citet{Li2023FindingSE} propose a diversity-guided example search method that iteratively refines and assesses chosen example permutations to identify the optimal permutation.
\citet{Scarlatos2023RetICLSR,Zhang2022ActiveES} regarded in-context example selection as a sequential decision-making problem and applied reinforcement learning to solve it. 
However, these studies are mainly focused on CausalLMs and do not delve into the causes behind order sensitivity.

\section{Conclusion}
In this paper, we discover that CausalLMs are more sensitive to the order of in-context examples compared to PrefixLMs likely due to the auto-regressive attention mask used in CausalLMs.
To mitigate this, we propose an unsupervised Information-Augmented and Consistency-Enhanced fine-tuning approach which employs contrastive learning to align representations of an in-context example across different positions with that at the end of the demonstration.
Additionally, a consistency loss is introduced to improve representation consistency near the prediction head by ensuring that representations from different permutations are similar.
Experimental results validate the efficacy of our proposed method across five benchmarks, highlighting notable cross-pool and cross-count generalizability.


%% file: sections/6_Appendix.tex
\section{Appendix / supplemental material}
\label{sec:appendix}

\renewcommand{\thefigure}{A\arabic{figure}}
\setcounter{figure}{0}

\renewcommand{\thetable}{A\arabic{table}}
\setcounter{table}{0}

\subsection{Overview of the Selected LLMs}
\label{LLMs}
Table~\ref{ModelInfo} provides details of the LLMs employed in our research, specifying the type, the max input length and the size of each LLM.
Throughout the paper, the term "Llama" is used to denote "Llama2-chat".
\begin{table*}[h]
\centering
\resizebox{0.7\linewidth}{!}{
\begin{tabular}{l|l|r|c}
	\toprule
	\textbf{Name} & \textbf{Type} & \textbf{Size} & \textbf{Max Input Length} \\
	\midrule
	Flan-T5-XL~\citep{Raffel2019ExploringTL} & PrefixLM &   3B & 2048 \\
	Flan-T5-XXL~\citep{Raffel2019ExploringTL} & PrefixLM &  11B & 2048  \\
	Flan-UL2~\citep{Tay2022UL2UL} & PrefixLM &  20B & 2048  \\
 \midrule
	Llama2-chat-7B~\citep{Touvron2023Llama2O} & CausalLM & 7B  & 4096     \\
	Llama2-chat-13B~\citep{Touvron2023Llama2O} & CausalLM & 13B & 4096     \\
	Vicuna-7B~\citep{Zheng2023JudgingLW} & CausalLM & 7B  & 4096    \\
	Vicuna-13B~\citep{Zheng2023JudgingLW} & CausalLM & 13B  & 4096    \\
	\bottomrule
\end{tabular}}
\caption{\label{ModelInfo}
	 Details about the models used in the experiments.}
\end{table*}

\subsection{Preliminary Experimental Results on SST-2 and MMLU Benchmarks with 4-shot}
\label{PreResults4shot}

\begin{table*}[h]
\centering
\resizebox{0.7\linewidth}{!}{
\begin{tabular}{l|ccc|ccc}
	\toprule
	Benchmark & \multicolumn{3}{c|}{SST-5} & \multicolumn{3}{c}{MMLU} \\
	\midrule
	Metrics & MV$(\uparrow)$ & Partial$(\downarrow)$ & Entropy$(\downarrow)$ & MV$(\uparrow)$ & Partial$(\downarrow)$ & Entropy$(\downarrow)$ \\
	\midrule
	Flan-T5-XL & 0.437  & 0.169 & 0.117 & 0.507 & 0.077 & 0.063     \\
	Flan-T5-XXL & 0.326 & 0.025 & 0.014 & 0.535 & 0.080 & 0.061     \\
	Flan-UL2 & 0.517 &  0.079 & 0.054   & 0.559 & 0.068 & 0.054     \\
 	\midrule
	Llama-7B  & 0.504 & 0.556 & 0.390 & 0.468 & 0.207 & 0.167 \\
	Llama-13B & 0.459 & 0.623 & 0.482 & 0.520 & 0.175 & 0.129 \\
	Vicuna-7B & 0.496 & 0.353 & 0.262 & 0.487 & 0.229 & 0.172 \\
	Vicuna-13B & 0.477 & 0.516 & 0.399 & 0.546 & 0.180 & 0.136 \\
	\bottomrule
\end{tabular}}
\caption{\label{PreExperimentResult4Shot}
	Experimental results on SST-5 and MMLU benchmarks with 4-shot.}
\end{table*}

Table~\ref{PreExperimentResult4Shot} represents experimental results on SST-5 and MMLU benchmarks with the number of in-context examples set to 4.
We use all 24 possible permutations to generate prompts and obtain their answers by feeding them to LLMs.
For the MMLU benchmark, we present the average experimental results for the same sub-tasks as shown in Table~\ref{PreExperimentResult10Shot}.  
We use the code released by \citep{Fu2023ChainofThoughtHA} to conduct the experiments.

\subsection{Similarities of the Last Layer Outputs of LLama After Fine-tuning with \textbf{InfoAC}}
\label{HeatInfoAC}
\begin{figure}[h]
 	\centering
 	\includegraphics[width=0.4\textwidth]{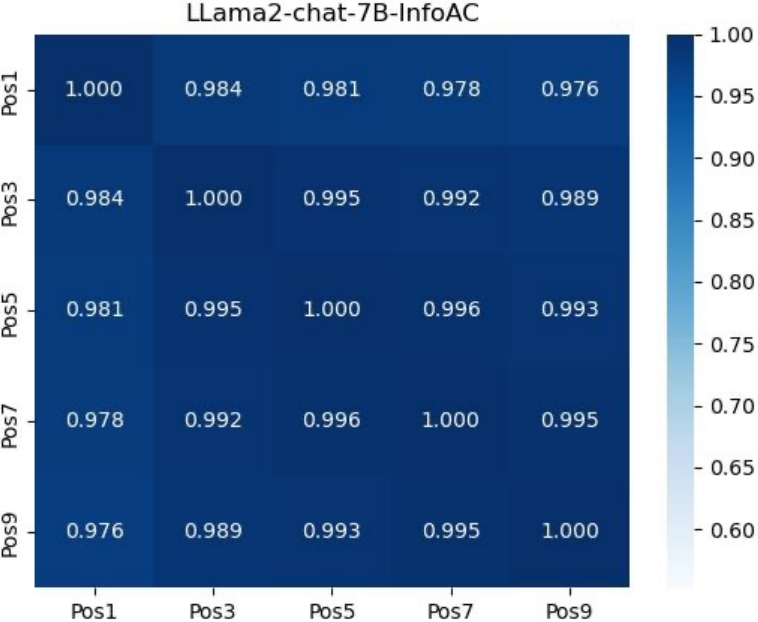} 
 	\caption{\label{heatLLamaInfoAC}
  The heatmap visualizes the similarities in representations of a specific token within a sample from the last layer outputs across different positions for Llama2-chat-7B after fine-tuning with \textbf{InfoAC}.
 	}
 \end{figure}

Figure~\ref{heatLLamaInfoAC} shows the heat map that visualises the similarities of representations of a sample across different positions for Llama after fine-tuning with InfoAC.
The chosen samples are identical to the ones featured in both Figure~\ref{heatLLama} and Figure~\ref{heatT5}.
 
\subsection{Statistics of Benchmarks}
\label{Benchmark}
Table~\ref{Datasets} provides an overview of the datasets used in our study, detailing the type of the benchmark and the size of the training, development, and test sets. Here is a brief introduction to the datasets:
\begin{itemize}
	\item \textbf{S}tanford \textbf{S}entiment \textbf{T}reebank with \textbf{5} Labels Dataset~\citep{Socher2013RecursiveDM}~(SST-5): This dataset consists of movie reviews annotated with five levels of sentiment: very positive, positive, neutral, negative, and very negative. 
	\item \textbf{S}tanford \textbf{S}entiment \textbf{T}reebank with \textbf{2} Labels Dataset~\citep{Socher2013RecursiveDM}~(SST-2): It is a binary sentiment analysis dataset, with two levels of sentiment: positive and negative.
	\item \textbf{Q}uora \textbf{Q}uestion \textbf{P}airs dataset~(QQP)~\citep{quora-question-pairs}: It is a dataset related to the problem of identifying duplicate questions. Each sample contains a pair of sentences, and labels of `duplicate' and `not duplicate' are assigned based on whether the semantic information of the two sentences overlaps.
	\item Sequence Next Term Dataset~\citep{2019arXiv}: This sub-task belongs to the Mathematics dataset~\citep{2019arXiv} and serves as a generation task designed to assess the mathematical capabilities of a language model (LM). Each sample consists of a sequence of numbers arranged in a specific pattern, and the task requires the LM to predict the next number in the sequence. 
	\item Round Number Dataset~\citep{2019arXiv}: It is also a generation task that forms part of the Mathematics dataset. Each sample's input is an instruction to round a decimal, and the answer is the result after operating as instructed.
\end{itemize}

\begin{table*}[h]
\centering
\resizebox{0.7\linewidth}{!}{
\begin{tabular}{l|c|r|r|r}
	\toprule
	Name & Type & Train & Dev & Test\\
	\midrule
	SST-5~\citep{Socher2013RecursiveDM} & Classification & 8.5k & 1.1k & 2.2k\\
	SST-2~\citep{Socher2013RecursiveDM} & Classification & 67.3k & 0.9k & 1.8k\\
	QQP~\citep{quora-question-pairs} & Classification & 364k & 40.4k & 391k   \\
	Sequence Next Term~\citep{2019arXiv} & Generation & 2M & - & 10k  \\
	Round Number~\citep{2019arXiv} & Generation & 2M & - & 10k   \\
	\bottomrule
\end{tabular}}
\caption{\label{Datasets}
	 Statistics regarding the benchmarks employed in the experiment.
}
\end{table*}

\subsection{Experimental Setup}
\label{ExperimentalSetup}
We adopt four CausalLMs including Llama2-chat-7B~\citep{Touvron2023Llama2O}, Llama2-chat-13B~\citep{Touvron2023Llama2O}, Vicuna-7B~\citep{Zheng2023JudgingLW} and Vicuna-13B~\citep{Zheng2023JudgingLW} as backbones. 
For all LLMs, the rank of LoRA is set to 8.
The hyperparameter $\lambda_1$ in the total loss function is set to 1.
These models are fine-tuned using the Adam optimizer~\cite{Kingma2014AdamAM} with a learning rate of 1e-4, and training is conducted for a single epoch. 
In the inference phase, we use a greedy search decoding algorithm to prevent the model from predicting different answers due to randomness.
Experiments are carried out on A100-80G GPUs, with most of the fine-tuning processes being completed within four hours.

For the settings of in-context learning, the number of in-context examples $k$ is set at 10. 
The template $T$ of our input consists of three parts: Instruction, demonstration and the test sample, which is shown as follows:
\begin{equation}
	\small
	T =  [\text{Instruction}, \text{Demonstration}, \text{Test}],
\end{equation}
\begin{equation}
	\small
	\text{Demonstration} = [``Q:\{x_i\} \ A:\{y_i\}"]_{\times10},
\end{equation}
\begin{equation}
	\small
	\text{Test} = ``Q:\{\text{Test Query}\} \ A:".
\end{equation}
Here, `Instruction' refers to the description of the task, and `Demonstration' consists of 10 in-context examples arranged in a certain permutation.
The query $x_i$ and answer $y_i$ of each in-context sample are positioned within the template following `Q:' and `A:', respectively.
`Test' refers to the test query we need the model to answer.
The default sizes for both the candidate pool $\mathbb{P}$ and the pseudo test set $\mathbb{P}_T$ are established at 100.  
We sample $N=1000$ training batches to construct the training dataset.
For each training batch, we sample $m=8$ permutations to construct prompts.
In the inference stage, 20 permutations are sampled for each test case to construct prompts, consistent with the preliminary experiments described in section~\ref{sec:pre_study}.

We evaluate our method on the development sets of the QQP and SST-2 datasets, which are part of the GLUE benchmark~\citep{wang2019glue}, owing to the unavailability of publicly released test labels. Specifically, for the QQP dataset, due to the large size of its original development set, we select 1000 development samples as our test set. 
In contrast, for the SST-2 benchmark, we perform our evaluation using its entire original development set.
The evaluation for the SST-5 dataset~\citep{Socher2013RecursiveDM} is carried out on its original test set. 
For the Sequence Next Term and Round Number datasets, we sample 1000 test samples for our evaluation. 

\begin{table*}[t]
\centering
\resizebox{0.5\linewidth}{!}{
\begin{tabular}{l|ccc}
	\toprule
	Dataset	& \multicolumn{3}{c}{SST-2}           \\
	\midrule
	Metrics & MV($\uparrow$)  & Partial($\downarrow$) & Entropy($\downarrow$)  \\
	\midrule
	Llama-7B & 0.934$\pm$0.005 & 0.081$\pm$0.005 & 0.050$\pm$0.004  \\
	+IA & 0.944$\pm$0.001 & 0.036$\pm$0.010 & 0.023$\pm$0.006  \\
	+InfoAC & \textbf{0.944}$\pm$\textbf{0.002} & \textbf{0.024}$\pm$\textbf{0.007} & \textbf{0.015}$\pm$\textbf{0.003} \\
	\midrule
	Llama-13B  & 0.935$\pm$0.004 & 0.080$\pm$0.015 & 0.050$\pm$0.009 \\
	+IA & 0.943$\pm$0.003 & 0.037$\pm$0.003 & 0.022$\pm$0.002 \\
	+InfoAC & \textbf{0.944}$\pm$\textbf{0.004} & \textbf{0.015}$\pm$\textbf{0.004} & \textbf{0.010}$\pm$\textbf{0.003} \\
	\midrule
	Vicuna-7B  & 0.938$\pm$0.007 & 0.074$\pm$0.004 & 0.045$\pm$0.003 \\
	+IA & 0.941$\pm$0.007 & 0.029$\pm$0.003 & 0.020$\pm$0.003 \\
	+InfoAC & \textbf{0.942}$\pm$\textbf{0.006} & \textbf{0.022}$\pm$\textbf{0.003} & \textbf{0.014}$\pm$\textbf{0.002} \\
	\midrule
	Vicuna-13B  & 0.924$\pm$0.009 & 0.065$\pm$0.011 & 0.040$\pm$0.006 \\
	+IA & 0.940$\pm$0.002 & 0.039$\pm$0.001 & 0.022$\pm$0.002 \\
	+InfoAC & \textbf{0.943}$\pm$\textbf{0.004} & \textbf{0.026}$\pm$\textbf{0.002} & \textbf{0.016}$\pm$\textbf{0.002}\\
	\bottomrule
\end{tabular}
}
\caption{\label{MainResults2}
	Experimental results on the SST-2 benchmark.
}
\end{table*}

\subsection{Implementation Details of the In-context Selection Approaches}
\label{InContextSelectionDetails}
In this section, we outline the implementation details of the two in-context selection approaches.
The setup of in-context learning is the same as the main experiments, as detailed in Appendix~\ref{ExperimentalSetup}.

For the KNN method, the candidate pool is randomly selected from the training set and consists of 100 examples.
For each test sample, we adopt RoBERTa~\citep{Liu2019RoBERTaAR} to embed the test query and all the candidate queries.
The average representation of all tokens in a sample is used as the sample's representation, and the semantic similarities between the test query and all candidate queries are calculated.
The 10 samples with the highest semantic similarity are selected as the in-context examples for the test case.

As for the One-Shot method, We reserve 100 samples from the training set to serve as $D_{dev}$.
The remaining 5000 training samples that do not appear in $D_{dev}$ are assessed to create the candidate pool.
The 100 samples with the highest one-shot in-context learning accuracy on the $D_{dev}$ are selected as the candidate pool.
For each test case, we randomly select 10 samples from this candidate pool as its in-context samples.

We construct training data for \textbf{InfoAC}  sharing the same candidate pool as both KNN and One-Shot respectively.
In the inference stage, \textbf{InfoAC} and the original LLM use exactly the same test dataset for evaluation.

\subsection{Generalizability Across Different Candidate Pools}
\label{CrossPool}

\begin{table*}[t]
\centering
\resizebox{0.9\linewidth}{!}{
\begin{tabular}{l|ccc|l|ccc}
	\toprule
	Metrics & MV$(\uparrow)$ & Partial$(\downarrow)$ & Entropy$(\downarrow)$ & Metrics & MV$(\uparrow)$ & Partial$(\downarrow)$ & Entropy$(\downarrow)$ \\
        \midrule
        \multicolumn{8}{c}{SST-5} \\
	\midrule
	Llama-7B & 0.509 & 0.559 & 0.391 & Vicuna-7B & \textbf{0.505} & 0.457 & 0.314 \\
    +\textbf{InfoAC}(Same) & \textbf{0.533} & 0.161 & 0.108 & +\textbf{InfoAC}(Same) & 0.497 & \textbf{0.108} & \textbf{0.066} \\
	+\textbf{InfoAC}(Diff)  & 0.520 & \textbf{0.160} & \textbf{0.108} & +\textbf{InfoAC}(Diff) & 0.503 & 0.138 & 0.095	  \\
	\midrule
	Llama-13B & 0.495 & 0.600 & 0.452 & Vicuna-13B & 0.490 & 0.497 & 0.356 \\
    +\textbf{InfoAC}(Same) & 0.502 & \textbf{0.176} & \textbf{0.117} & +\textbf{InfoAC}(Same) & \textbf{0.520} & 0.224 & 0.149\\ 
	+\textbf{InfoAC}(Diff) & \textbf{0.509} & 0.188 & 0.120 & +\textbf{InfoAC}(Diff) & 0.517 & \textbf{0.195} & \textbf{0.128}     \\
        \midrule
        \multicolumn{8}{c}{Sequence Next Term} \\
        \midrule
        Llama-7B & 0.233 & 0.369 & 0.342 & Vicuna-7B & 0.226 & 0.504 & 0.546 \\
    +\textbf{InfoAC}(Same) & \textbf{0.257} & \textbf{0.175} & 0.147 & +\textbf{InfoAC}(Same) & \textbf{0.234} & \textbf{0.235} & 0.198 \\
	+\textbf{InfoAC}(Diff)  & 0.244 & 0.189 & \textbf{0.132} & +\textbf{InfoAC}(Diff) & 0.219 & 0.237 & \textbf{0.184}	  \\
	\midrule
	Llama-13B & 0.230 & 0.404 & 0.372 & Vicuna-13B & 0.233 & 0.442 & 0.431 \\
    +\textbf{InfoAC}(Same) & 0.284 & \textbf{0.134} & \textbf{0.108} & +\textbf{InfoAC}(Same) & 0.278 & \textbf{0.187} & \textbf{0.159}\\ 
	+\textbf{InfoAC}(Diff) & \textbf{0.289} & 0.166 & 0.121 & +\textbf{InfoAC}(Diff) & \textbf{0.281} & 0.242 & 0.181    \\
	\bottomrule
\end{tabular}}
\caption{\label{CrossPoolSST5}
	Cross-Pool experimental results on the SST-5 and Sequence Next Term dataset. ``Diff'' denotes that the training set for \textbf{InfoAC} from a candidate pool differs from the candidate pool for the test data. ``Same'' denotes that the same pool is used for both the training and testing phases.}
\end{table*}

In this section, we explore the generalizability of the proposed \textbf{InfoAC} when the demonstrations used during inference are not from the same candidate pool in training.
Table~\ref{CrossPoolSST5} presents the results for the SST-5 and the Sequence Next Term dataset benchmark.
It shows that our method has good generalizability to different candidate pools. 
When a different candidate pool is used during testing, the model still significantly reduces its sensitivity to the order of in-context examples.
The main reason for this is that our method is trained in an unsupervised manner, aiming to learn a better representation for each in-context example. 
Since the information in various samples shares similarities, the insights and knowledge acquired by the model from one candidate set can also help reduce the order sensitivity of in-context examples in another candidate set.

\subsection{Additional Supplementary Experimental Results on the Sequence Next Term benchmark}
\label{SupplExperiment}
In this section, we provide in-context selection and cross-count results on the Sequence Next Term benchmark, as detailed in Table~\ref{InContextSelectionNext}, and Table~\ref{CrossLengthNext}.
As shown in Table~\ref{InContextSelectionNext}, using KNN for in-context selection on the Sequence Next Term dataset does not lead to performance improvement when compared to random selection~\footnote{Results can be found in Table~\ref{MainResults1}}.
This is because the query of this dataset consists of a sequence of numbers and does not contain semantic information.
KNN selects samples that are semantically most similar to the test case as in-context examples, which is not suitable for this dataset.

\begin{table*}[t]
\centering
\resizebox{0.7\linewidth}{!}{
\begin{tabular}{l|ccc|ccc}
	\toprule
	Metrics & MV$(\uparrow)$  & Partial$(\downarrow)$ & Entropy$(\downarrow)$ & MV$(\uparrow)$  & Partial$(\downarrow)$ & Entropy$(\downarrow)$ \\
	\midrule
	& \multicolumn{3}{c|}{KNN} & \multicolumn{3}{c}{One-Shot} \\ 
	\midrule
	Llama-7B & 0.212 & 0.458 & 0.502 & 0.270 & 0.359 & 0.354 \\
	+\textbf{InfoAC}  & \textbf{0.255} & \textbf{0.204} & \textbf{0.177} & \textbf{0.277} & \textbf{0.181} & \textbf{0.141}	  \\
	\midrule
	Llama-13B & 0.262 & 0.447 & 0.431 & 0.299 & 0.314 & 0.250 \\
	+\textbf{InfoAC} & \textbf{0.283} & \textbf{0.170} & \textbf{0.140} & \textbf{0.312} & \textbf{0.151} & \textbf{0.112} \\
	\midrule
	Vicuna-7B & 0.202 & 0.535 & 0.604 & 0.256 & 0.473 & 0.462  \\
	+\textbf{InfoAC} & \textbf{0.234} & \textbf{0.252} & \textbf{0.232} & \textbf{0.272} & \textbf{0.206} & \textbf{0.154} \\
	\midrule
	Vicuna-13B & 0.276 & 0.460 & 0.467 & 0.290 & 0.303 & 0.280  \\
	+\textbf{InfoAC} & \textbf{0.283} & \textbf{0.208} & \textbf{0.163} & \textbf{0.294} & \textbf{0.228} & \textbf{0.158}\\
	\bottomrule
\end{tabular}}
\caption{\label{InContextSelectionNext}
	Performances of different in-context sample selection methods on the  Sequence Next Term dataset.
	}
\end{table*}

\begin{table*}[t]
\centering
\resizebox{\linewidth}{!}{
\begin{tabular}{l|ccc|ccc|ccc}
    \toprule
    Metrics & MV$(\uparrow)$ & 
    Partial$(\downarrow)$ & Entropy$(\downarrow)$ & MV$(\uparrow)$ & 
    Partial$(\downarrow)$ & Entropy$(\downarrow)$ & MV$(\uparrow)$  & Partial$(\downarrow)$ & Entropy$(\downarrow)$\\
    \midrule
    & \multicolumn{3}{c|}{k=4} & \multicolumn{3}{c|}{k=20} & \multicolumn{3}{c}{k=50} \\
    \midrule
    Llama-7B & 0.233 & 0.283 & 0.261 & 0.251 & 0.482 & 0.452 & 0.229 & 0.550 & 0.583\\
    +\textbf{InfoAC}  & \textbf{0.263} & \textbf{0.163} & \textbf{0.123} & 	 \textbf{0.266} & \textbf{0.214} & \textbf{0.187} & \textbf{0.252} & \textbf{0.337} & \textbf{0.274} \\
    \midrule
    Llama-13B & 0.261 & 0.264 & 0.232 & 0.287 & 0.380 & 0.316 & \textbf{0.287} & 0.390 & 0.331 \\
    +\textbf{InfoAC} & \textbf{0.286} & \textbf{0.119} & \textbf{0.092} & \textbf{0.287} & \textbf{0.164} & \textbf{0.142}  & 0.281 & \textbf{0.178} & \textbf{0.141} \\
    \midrule
    Vicuna-7B & 0.210 & 0.424 & 0.436 & 0.229 & 0.546 & 0.583 & 0.237 & 0.561 & 0.700 \\
    +\textbf{InfoAC} & \textbf{0.244} & \textbf{0.238} & \textbf{0.191} & \textbf{0.236} & \textbf{0.292} & \textbf{0.257} & \textbf{0.258} & \textbf{0.411} & \textbf{0.417} \\
    \midrule
    Vicuna-13B & 0.271 & 0.369 & 0.344 & 0.283 & 0.378 & 0.327 & \textbf{0.277} & 0.343 & 0.330\\
    +\textbf{InfoAC} & \textbf{0.284} & \textbf{0.211} & \textbf{0.161} & \textbf{0.288} & \textbf{0.236} & \textbf{0.199} & 0.271 & \textbf{0.173} & \textbf{0.155}\\
    \bottomrule
\end{tabular}}
\caption{\label{CrossLengthNext}
	Cross-count experimental results on the Sequence Next Term dataset.}
\end{table*}

\subsection{Ethical Statement}
In this paper, we conduct preliminary experiments on the Massive Multitask Language Understanding~\citep{Hendrycks2020MeasuringMM} (MMLU) benchmark to explore the impact of the order of in-context examples on the performance of PrefixLMs and CausalLMs.
This benchmark is crafted to assess the knowledge acquisition of LLMs via pretraining, by evaluating them exclusively in zero-shot and few-shot settings. It encompasses 57 sub-tasks, covering a wide range of areas including STEM, the humanities, and beyond.
Certain sub-tasks, like ``moral scenarios'' and ``moral disputes'', might include statements that elicit ethical considerations.
There is a certain risk that LLMs may misuse text from datasets, therefore, we recommend thoroughly considering safety before practical application.